\pdfoutput=1

\documentclass[11pt]{article}

\usepackage[preprint]{coling}

\usepackage{times}
\usepackage{latexsym}

\usepackage[T1]{fontenc}

\usepackage[utf8]{inputenc}

\usepackage{microtype}

\usepackage{inconsolata}

\usepackage{graphicx}

\usepackage[stable]{footmisc}
\usepackage[greek, english]{babel}
\newcommand{\greeknlptoolkit}{\textsc{gr-nlp-toolkit}\xspace}

\newcommand\nlp{\textsc{nlp}\xspace}
\newcommand\api{\textsc{api}\xspace}
\newcommand\pos{\textsc{pos}\xspace}
\newcommand\ner{\textsc{ner}\xspace}
\newcommand\pypi{\textsc{pypi}\xspace}
\newcommand{\greeknlpapi}{\textsc{greek-nlp-api}\xspace}
\newcommand\httpapi{\textsc{http api}\xspace}
\newcommand\stanza{\textsc{stanza}\xspace}
\newcommand\spacy{\textsc{spacy}\xspace}
\newcommand\xlmr{\textsc{xlm-r}\xspace}
\newcommand\greekbert{\textsc{greek-bert}\xspace}
\newcommand\meltemi{\textsc{meltemi}\xspace}
\newcommand\mistral{\textsc{mistral}\xspace}
\newcommand\llm{\textsc{llm}\xspace}
\newcommand\bytfive{\textsc{byt5}\xspace}
\newcommand\nltk{\textsc{nltk}\xspace}
\newcommand\fasttext{\textsc{fasttext}\xspace}
\newcommand\dparsing{\textsc{dp}\xspace}
\newcommand\gtog{\textsc{g2g}\xspace}
\newcommand\gptfour{\textsc{gpt-4}\xspace}

\usepackage{xspace}
\usepackage{mciteplus}
\usepackage[T1]{fontenc}

\definecolor{darkerred}{RGB}{200,0,0}
\definecolor{darkergreen}{RGB}{0,150,0}

\usepackage{pifont}
\usepackage{amsmath}

\title{\greeknlptoolkit: An Open-Source NLP Toolkit for Modern Greek}

\author{Lefteris Loukas\textsuperscript{1,3}, Nikolaos Smyrnioudis\textsuperscript{1}, Chrysa Dikonomaki\textsuperscript{1}, Spyros Barbakos\textsuperscript{1},\\
\textbf{Anastasios Toumazatos\textsuperscript{1}}, \textbf{John Koutsikakis\textsuperscript{1}}, \textbf{Manolis Kyriakakis\textsuperscript{1}}, \textbf{Mary Georgiou\textsuperscript{1}},\\
\textbf{Stavros Vassos\textsuperscript{3}}, \textbf{John Pavlopoulos\textsuperscript{1,2}}, \textbf{Ion Androutsopoulos\textsuperscript{1,2}} \\
        \textsuperscript{1}Department of Informatics, Athens University of Economics and Business, Greece\\
        \textsuperscript{2}Archimedes/Athena RC, Greece\\
        \textsuperscript{3}helvia.ai}

\begin{document}
\maketitle
\begin{abstract}
We present \greeknlptoolkit, an open-source natural language processing (\nlp) toolkit developed specifically for modern Greek. The toolkit provides state-of-the-art performance in five core \nlp tasks, namely part-of-speech tagging, morphological tagging, dependency parsing, named entity recognition, and Greeklish-to-Greek transliteration. The toolkit is based on pre-trained Transformers, it is freely available, and can be easily installed in Python \mbox{(\texttt{pip install gr-nlp-toolkit})}. It is also accessible through a demonstration platform on HuggingFace, along with a publicly available \api for non-commercial use. We discuss the functionality provided for each task, the underlying methods, experiments against comparable open-source toolkits, and future possible enhancements. The toolkit is available at: \url{https://github.com/nlpaueb/gr-nlp-toolkit}

\end{abstract}

\section{Introduction}\label{sec:introduction}

Modern Greek is the official language of Greece, one of the two official languages of Cyprus, and the native language of approximately 13 million people.\footnote{\url{https://en.wikipedia.org/wiki/Greek_language}} Despite 
continuous efforts 
\cite{Papantoniou2020,bakagianni2024towards}, there are still very few 
natural language processing (\nlp) toolkits that support modern Greek (\S\ref{sec:related_work}).

We present \greeknlptoolkit, an open-source \nlp toolkit developed specifically for modern Greek. The toolkit supports five core \nlp tasks, namely part-of-speech (\pos) tagging, morphological tagging (tagging for tense, voice, person, gender, case, number etc.), dependency parsing, named entity recognition (\ner), and Greeklish-to-Greek transliteration (converting Greek written using Latin-keyboard characters to the Greek alphabet). We demonstrate the functionality that the toolkit provides per task (\S\ref{sec:the_toolkit}). We also discuss the underlying methods and experimentally compare \greeknlptoolkit to \stanza \cite{qi-etal-2020-stanza} and \spacy \cite{honnibal2020spacy}, two multilingual toolkits that support modern Greek, demonstrating that \greeknlptoolkit achieves state-of-the-art performance in \pos tagging, morphological tagging, dependency parsing, and \ner (\S\ref{sec:models_and_methods}). Previous work \cite{toumazatos-etal-2024-still-all-greeklish-to-me} shows that the Greeklish-to-Greek converter included in \greeknlptoolkit is also state-of-the-art. 

The toolkit can be easily installed in Python via \pypi (\mbox{\texttt{pip install gr-nlp-toolkit}}) and its code is publicly available on Github.\footnote{\url{https://github.com/nlpaueb/gr-nlp-toolkit/}} We showcase its functionality in an open-access demonstration space, hosted on HuggingFace.\footnote{\url{https://huggingface.co/spaces/AUEB-NLP/greek-nlp-toolkit-demo}} We also release \greeknlpapi, a fully-documented and publicly available \httpapi, which allows using the toolkit in (non-commercial) applications developed in any programming language.\footnote{\url{https://huggingface.co/spaces/AUEB-NLP/The-Greek-NLP-API/}}

\section{Background and related work}\label{sec:related_work}

Greek has evolved 
over three millennia.\footnote{\url{www.britannica.com/topic/Greek-language}} 
Apart from its historical interest, Greek is also challenging from an \nlp point of view. 
For example, it has its own alphabet (\textgreek{α,β,γ,..}), and nowadays a much smaller number of speakers, compared to other widely used languages of the modern world. Although words of Greek origin can be found in many other languages (e.g., medical terms), they are written in different alphabets in other languages. Hence, Greek words written in the Greek alphabet are severely under-represented in modern multilingual corpora and, consequently, in the word and sub-word vocabularies of most multilingual Transformer models, e.g., \xlmr \cite{conneau-etal-2020-unsupervised}. This causes the tokenizers of these models to over-fragment Greek words, very often to characters \cite{greekbert20020}, which increases processing time and cost, and 
makes it more difficult for models to reassemble tokens to more meaningful units.  
Greek is also highly inflected (e.g., different verb forms for different tenses, voices, moods, persons, numbers; similarly for nouns, adjectives, pronouns etc.), which makes \pos tagging more difficult and morphological tagging (tagging also for tense, voice, gender, case etc.) desirable. Greek is also flexible in word order (e.g., subject-verb-object, object-verb-subject, verb-subject-object etc.\ are all possible with different emphasis), which makes parsing more challenging. 

\begin{figure}[h]
  \includegraphics[width=\columnwidth]{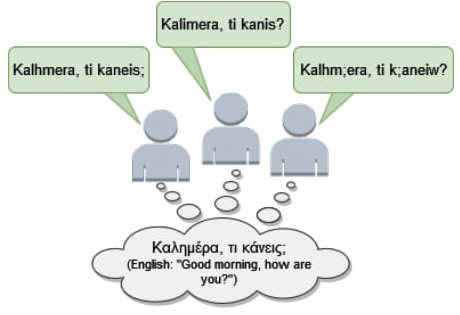}
  \caption{An example of a Greek sentence written in Greeklish. There is no consensus mapping. Greek characters may be replaced by Latin-keyboard characters based on visual similarity, phonetic similarity, shared keys etc. Figure from \citet{toumazatos-etal-2024-still-all-greeklish-to-me}.}
  \label{fig:g2g_example}
\end{figure}

Modern Greek is normally written in the Greek alphabet.
In online messages, however, especially informal email and chat, it is often written using characters available on Latin-character keyboards, a form known as Greeklish \cite{koutsogiannis-greeklish-2017}. For example, `$\omega$' (omega) may be written as `w' based on visual similarity, as `o' based on phonetic similarity, or as `v' based on the fact that `$\omega$' and `v' use the same key on Greek-Latin keyboards, to mention just some possibilities. Greeklish was originally used in older computers that did not support the Greek alphabet, but continues to be used to avoid switching languages on multilingual keyboards, hide spelling mistakes (esp.\ when used by non-native speakers), or as a form of slang (mostly by younger people). There is no consensus mapping between Greek and Latin-keyboard characters.\footnote{The \textsc{iso} 843:1997 standard (\url{https://www.iso.org/standard/5215.html}) is almost never used.} Consequently, the same Greek word can be written in numerous different ways in Greeklish (Fig.~\ref{fig:g2g_example}). Even native Greek speakers may struggle to understand, and are often annoyed by Greeklish, which requires paying careful attention to context to decipher. Moreover, most Greek \nlp datasets contain text written in the Greek alphabet, hence models trained on those datasets may be unable to handle Greeklish.  

Phenomena of this kind motivated the development of \greekbert \cite{greekbert20020}, and more recently the \meltemi large language model (\llm) for modern Greek \cite{voukoutis2024meltemiopenlargelanguage}; the latter is based on \mistral-7b \cite{jiang2023mistral7b}. In this work, we leverage \greekbert for most tasks, and \bytfive \cite{xue-etal-2022-byt5} for Greeklish-to-Greek, which can both be used even with \textsc{cpu} only, unlike larger \llm{s} \cite{power-hungry-processing-24}.
Nevertheless, in future versions of the toolkit, we plan to investigate how we can integrate `small' Greek \llm{s} for on-device use.\footnote{For example, Meta recently released 1B and 3B \llm{s} for on-device use: \url{https://ai.meta.com/blog/llama-3-2-connect-2024-vision-edge-mobile-devices/}}

In previous modern Greek experiments, \greekbert, when fine-tuned, was reported to outperform the multilingual \xlmr, again fine-tuned, in \ner and natural language inference, while it performed on par with \xlmr in \pos tagging \cite{greekbert20020}. 
In subsequent work of two undergraduate theses \cite{Dikonimaki2021,Smyrnioudis2021}, we showed, again using modern Greek data, that \greekbert largely outperformed \xlmr in dependency parsing, but found no substantial difference between the two models in morphological tagging and (another dataset of) \ner. Greeklish was not considered in any of these previous studies. The two theses also created a first version of \greeknlptoolkit, which was largely experimental, did not include Greeklish-to-Greek, and was not published (apart from 
the two theses). The version of the toolkit that we introduce here has been completely refactored, it uses more recent libraries, has been tested more thoroughly, includes Greeklish-to-Greek, can be used via both \pypi and \greeknlpapi,  and can also be explored via a HuggingFace demo (\S\ref{sec:introduction}).

\begin{table*}[t]
\centering
\small 
\label{tab:comparison}
\begin{tabular}{|c|c|c|c|c|c|c|}
\hline
\textbf{Toolkit} & \textbf{\begin{tabular}[c]{@{}c@{}}POS\\Tagging\end{tabular}} & 
\textbf{\begin{tabular}[c]{@{}c@{}}Morphological\\Tagging\end{tabular}} & 
\textbf{\begin{tabular}[c]{@{}c@{}}Lemma-\\tization\end{tabular}} 
& \textbf{\begin{tabular}[c]{@{}c@{}}Named Entity\\Recognition\end{tabular}} & \textbf{\begin{tabular}[c]{@{}c@{}}Dependency\\Parsing\end{tabular}} & \textbf{\begin{tabular}[c]{@{}c@{}}Greeklish-to-Greek\\Transliteration\end{tabular}} \\
\hline
\spacy & \textcolor{darkergreen}{\ding{51}} & \textcolor{darkergreen}{\ding{51}} & \textcolor{darkergreen}{\ding{51}} & \textcolor{darkergreen}{\ding{51}} & \textcolor{darkergreen}{\ding{51}} & \textcolor{darkerred}{\ding{55}} \\
\hline
\stanza & \textcolor{darkergreen}{\ding{51}} &\textcolor{darkergreen}{\ding{51}} & \textcolor{darkergreen}{\ding{51}} & \textcolor{darkerred}{\ding{55}} & \textcolor{darkergreen}{\ding{51}} & \textcolor{darkerred}{\ding{55}} \\
\hline
\nltk & \textcolor{darkerred}{\ding{55}} & \textcolor{darkerred}{\ding{55}} & \textcolor{darkerred}{\ding{55}} & \textcolor{darkerred}{\ding{55}} & \textcolor{darkerred}{\ding{55}} & \textcolor{darkerred}{\ding{55}} \\
\hline
\textbf{\greeknlptoolkit} & \textcolor{darkergreen}{\ding{51}} \textcolor{darkergreen}{\ding{51}} & \textcolor{darkergreen}{\ding{51}} \textcolor{darkergreen}{\ding{51}} & 
\textcolor{darkerred}{\ding{55}} &\textcolor{darkergreen}{\ding{51}} \textcolor{darkergreen}{\ding{51}} & \textcolor{darkergreen}{\ding{51}} \textcolor{darkergreen}{\ding{51}} & \textcolor{darkergreen}{\ding{51}} \textcolor{darkergreen}{\ding{51}} \\
\hline
\end{tabular}
\caption{Comparison of \nlp toolkits that support modern Greek. \nltk provides only a tokenizer and stop-word removal (not shown) for modern Greek. \spacy and \stanza both include a Greek lemmatizer. \greeknlptoolkit is the only one that includes Greeklish-to-Greek transliteration. Its other functions (\pos tagging, morphological tagging, \ner, dependency parsing) are based on \greekbert, whereas \spacy and \stanza are based on Greek \fasttext embeddings and do not use Transformers. \textcolor{darkergreen}{\ding{51}}\textcolor{darkergreen}{\ding{51}} denotes using pretrained Transformers.}
\label{tab:related_nlp_toolkits}
\end{table*}

\spacy \cite{honnibal2020spacy} and \stanza \cite{qi-etal-2020-stanza} are 
widely used multilingual \nlp toolkits that support modern Greek. They both have limitations, however, discussed below (Table \ref{tab:related_nlp_toolkits}).  

\spacy \cite{honnibal2020spacy} is an open-source \nlp library for efficient processing of text in many languages.
In modern Greek, it supports \pos tagging, morphological tagging, lemmatization (mapping all inflected forms of verbs, nouns etc.\ to their base forms), \ner, and dependency parsing. However, it 
relies on static Greek \fasttext word embeddings \cite{greek-ud-treebank-1-prokopidis-papageorgiou:2017:UDW,bojanowski-etal-2017-enriching}, without utilizing pretrained Transformers for modern Greek, which restricts its performance Also, \spacy does not support Greeklish-to-Greek transliteration.

Stanford's \stanza \cite{qi-etal-2020-stanza} is a Python-based \nlp library with multilingual support. For modern Greek, it provides \pos tagging, morphological tagging, dependency parsing, lemmatization, but not \ner or Greeklish-to-Greek. Its modern Greek components are 
trained on two Greek Universal Dependencies treebanks, the default `\textsc{gdt}' \cite{greek-ud-treebank-1-prokopidis-papageorgiou:2017:UDW, greek-ud-treebank-2-prokopidis-papageorgiou:2014:SPMRL-SANCL, PDKPP_2005,PDKPP_2006, GDKPP_2007}
and `\textsc{gud}' \cite{UD_Greek_GUD-markantonatou}. Under the hood, \stanza and \spacy use the same Greek \fasttext embeddings \cite{bojanowski-etal-2017-enriching} and no pretrained Transformers.

Another widely used \nlp toolkit, \nltk \cite{bird2009natural-nltk}, does not provide any functionality for modern Greek, other than a tokenizer and stop-word removal. In other related work, \citet{neural-resources-for-greek-nlp-prokopidis-and-piperidis-2020} introduced models for Greek \pos tagging, lemmatization, dependency parsing, and text classification, requiring manual integration with \fasttext and an outdated \stanza version. They also developed a closed-source \api based on them. 
By contrast, we focus on ready-to-use open-source \nlp toolkits.

\section{Using \greeknlptoolkit}\label{sec:the_toolkit}

\begin{figure}[h]
  \includegraphics[width=\columnwidth]{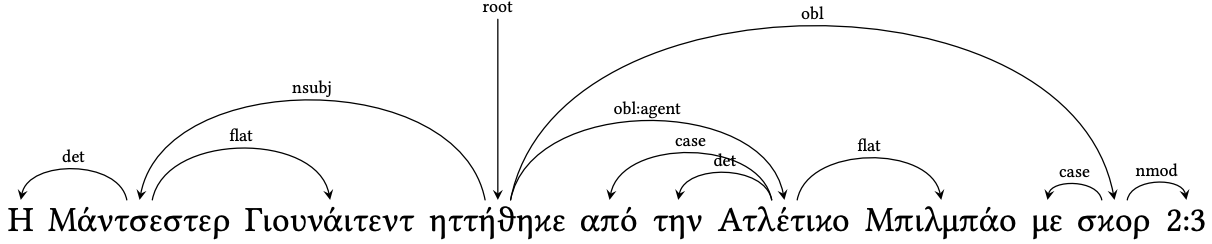}
  \caption{A dependency tree generated by \greeknlptoolkit for a Greek sentence whose English translation is ``Manchester United was defeated by Atletico Bilbao with a 2:3 score.'' Figure from \citet{Smyrnioudis2021}. Tree drawn using \spacy's visualizer.}
  \label{fig:dependency_parsing_figure}
\end{figure}

Using our toolkit in Python is straightforward. To install it, use \mbox{\texttt{pip install gr-nlp-toolkit}}. Subsequently, you can initialize, e.g., a pipeline for \pos tagging (incl.\  morphological tagging), \ner, dependency parsing (\dparsing) by executing \mbox{\texttt{nlp = Pipeline("pos, ner, dp")}}. Applying the pipeline to 
a sentence, e.g., \texttt{doc = nlp(``\textgreek{Η Ιταλία κέρδισε την Αγγλία στον τελικό το 2020.}'')}, tokenizes the text and provides linguistic annotations, including \pos and morphological tags, \ner labels, and dependency relations. In our  example, the token \textgreek{\textit{``Ιταλία"}} (English: `Italy') gets the annotations \texttt{NER = S-ORG} (start token of organization name), \texttt{UPOS = PROPN} (proper name), and a dependency relation \texttt{nsubj} (nominal subject) linking it to the verb (see also Fig. \ref{fig:dependency_parsing_figure}).

Transliterating Greeklish to Greek (\gtog) is equally simple. The \gtog converter can be loaded by typing \texttt{nlp = Pipeline("g2g")}. 
Running \texttt{doc = nlp("h athina kai h thessaloniki einai poleis")} will convert the text to ``\textgreek{\textit{η αθηνα και η θεσσαλονικη ειναι πολεις}}'' (English: ``athens and thessaloniki are cities''). 
This makes it easy to process Greeklish text before performing further Greek language processing. For example, you can also combine the \gtog converter with \pos, \ner, \dparsing in the same pipeline, using \mbox{\texttt{nlp = Pipeline("g2g, pos, ner, dp")}}.


\section{Under the hood and experiments}
\label{sec:models_and_methods}

        
The \pos tagging, morphological tagging, \ner, and dependency parsing tools of \greeknlptoolkit
are powered by \greekbert \cite{greekbert20020}, with task-specific heads.\footnote{\greekbert works as our backbone model in most tasks. While it is powerful, one limitation is that it automatically converts all text to lowercase and removes Greek accents.}
For Greeklish-to-Greek, we reproduced the \bytfive-based converter of citet{toumazatos-etal-2024-still-all-greeklish-to-me}, which was the best among several methods considered, apart from \mbox{\gptfour}, which we excluded for efficiency reasons.\footnote{\llm{s} like \gptfour or the Greek \meltemi require a significant resources (cost, time, lots of \textsc{vram}), which typical end users do not have.}

\subsection{Named entity recognition}\label{ner_subsection}

For the \ner tool of \greeknlptoolkit, we fine-tuned \greekbert \cite{greekbert20020} with a task-specific token classification head. We used the training subset of a modern Greek \ner dataset published by \citet{bartziokas2020datasets-elner}. The dataset contains approx.\ 38,000 tagged 
entities and 18 entity types.\footnote{The 18 entity types of \greeknlptoolkit are: \texttt{ORG, PERSON, CARDINAL, GPE, DATE, PERCENT, ORDINAL, LOC, NORP, TIME, MONEY, EVENT, PRODUCT, WORK\_OF\_ART, FAC, QUANTITY, LAW, LANGUAGE.}}
We tuned hyper-parameters to maximize the macro-F1 score on the development subset. We used cross-entropy loss, AdamW \cite{loshchilov2017fixing}, and grid search for hyper-parameter tuning (Table \ref{tab:greekbert_based_features_hyperparameters}).

In Table~\ref{tab:model_comparison_reduced}, we compare \spacy
against \greeknlptoolkit on the test subset of the \ner dataset of \citet{bartziokas2020datasets-elner}, for the six entity types that \spacy supports.\footnote{We provide the results only about the six shared \ner entity types between \spacy and \greeknlptoolkit.} We do not compare against \stanza here, since it does not support \ner (Table~\ref{tab:related_nlp_toolkits}).
As seen in Table~\ref{tab:model_comparison_reduced}, \greeknlptoolkit
outperforms \spacy in all entity types.\footnote{Table~\ref{tab:model_comparison_reduced} shows results of \spacy's large model (spaCy-lg). The smaller models (spacy-sm, spacy-md) performed worse.} 
\spacy's score in the \textsc{loc} (location) entity type is particularly low, because it classified most (truly) \textsc{loc} entities as \textsc{gpe} (geo-political entity).

\begin{table}[h!]
\small 
\centering
\begin{tabular}{l|c|c}
\hline 
\textbf{Entity type} & \textbf{\spacy} & \textbf{\greeknlptoolkit} \\
\hline
EVENT   & 0.31 & \textbf{0.64} \\
GPE     & 0.77 & \textbf{0.93} \\
PERSON  & 0.82 & \textbf{0.96} \\
LOC     & 0.01 & \textbf{0.80} \\
ORG     & 0.65 & \textbf{0.88} \\
PRODUCT & 0.27 & \textbf{0.75} \\ \hline

\end{tabular}
\caption{F1 test scores of \spacy and \greeknlptoolkit in modern Greek \ner, showed for the six entity types that \spacy supports.}
\label{tab:model_comparison_reduced}
\end{table}\vspace{-4mm}

\subsection{\pos tagging and morphological tagging} 
\label{subsec:pos_tagging}

For \pos tagging and morphological tagging, we used 
the modern Greek part of the Universal Dependencies (\textsc{ud}) treebank \cite{greek-ud-treebank-1-prokopidis-papageorgiou:2017:UDW}.
Every word occurrence is annotated with its gold universal \pos tag (\textsc{upos}), morphological features (\textsc{feats}), as well as its syntactic head and the type of syntactic dependency.
We refer the reader to the \textsc{ud} website, where complete lists of \textsc{upos} tags, morphological features, and dependency types are available.\footnote{\url{https://universaldependencies.org/}}

We fine-tuned a single \greekbert instance for both \pos tagging and morphological tagging, adding 17 token classification heads (linear layers), 16 for the morphological categories, and 1 additional token classification head for \textsc{upos} prediction.  Each classification head takes as input the corresponding output (top-level) token embedding of \greekbert. For every head, the class with the highest logit is chosen, as in multi-task learning. The model hyperparameters were tuned on the validation subset of the dataset optimizing the macro-F1 score, using grid search and AdamW \cite{loshchilov2017fixing} (Table \ref{tab:greekbert_based_features_hyperparameters}). 

In Table~\ref{tab:upos_tagging}, we compare \spacy and \stanza to  the \greeknlptoolkit on the \textsc{upos} and morphological tagging test data of the modern Greek \textsc{ud} treebank. \stanza and \greeknlptoolkit perform on par, with \spacy ranking third.

\begin{table}[h!]
\small
\centering
\begin{tabular}{l|ccc}
\hline
\textbf{Metric} & \textbf{\spacy} & \textbf{\stanza} & \textbf{\greeknlptoolkit} \\ \hline
Micro-F1        & 0.95        & \textbf{0.98}        & \textbf{0.98}      \\ \hline
Macro-F1        & 0.87        & 0.96            & \textbf{0.97}           \\ \hline
\end{tabular}
\caption{Micro-F1 and macro-F1 test scores for \textsc{upos} tagging. The complete list of \textsc{upos} tags can be found in \url{https://universaldependencies.org/u/pos/}.}
\label{tab:upos_tagging}
\end{table}


In the more complex morphological tagging task (Table \ref{tab:morphological_tagging}), the differences between the systems are move visible, with \greeknlptoolkit performing slightly better in most categories than \stanza, while \spacy, again, ranks third. The largest differences are observed in `Mood' and `Foreign' (foreign word), where \greeknlptoolkit performs substantially better, and `Degree' (degrees of adjectives), where \stanza is clearly better. \citet{Dikonimaki2021} attributes some of these differences to  very few training occurrences of the corresponding tags.

\begin{table}[h]
\scriptsize
\centering
\begin{tabular}{l|c|c|c}
\hline
\textbf{Morphological tag} &  \spacy & \stanza & \textbf{\greeknlptoolkit} \\
\hline
Case & 0.68 & \textbf{0.97} & \textbf{0.97} \\
Definite & 0.89 & \textbf{1.00} & \textbf{1.00} \\
Gender & 0.68 & 0.97 & \textbf{0.98} \\
Number & 0.69 & \textbf{0.99} & \textbf{0.99} \\
PronType & 0.71 & 0.94 & \textbf{0.97} \\
Foreign & 0.65 & 0.79 & \textbf{0.88} \\
Aspect & 0.65 & 0.98 & \textbf{0.99} \\
Mood & 0.74 & 0.59 & \textbf{0.83} \\
Person & 0.68 & 0.98 & \textbf{1.00} \\
Tense & 0.76 & 0.98 & \textbf{1.00} \\
VerbForm & 0.65 & \textbf{0.97} & 0.93 \\
Voice & 0.65 & \textbf{0.99} & 0.96 \\
NumType & 0.67 & 0.93 & \textbf{0.96} \\
Poss & 0.59 & 0.96 & \textbf{0.98} \\
Degree & 0.48 & \textbf{0.89} & 0.50 \\
Abbr & 0.89 & \textbf{0.96} & 0.94 \\
\hline
\end{tabular}
\caption{F1 test scores for all of the morphological tags.}
\label{tab:morphological_tagging}
\end{table}

\subsection{Dependency parsing}
\label{subsec:dependency_parsing}

For dependency parsing, we use the model of \citet{dozat-etal-2017-stanfords}, with the exception that we obtain contextualized word embeddings using \greekbert instead of the \textsc{bilstm} encoder of the original model.\footnote{When a word is broken into multiple sub-word tokens by \greekbert's tokenizer, we take the embedding of the first token to represent the entire word.} Specifically, for each word of the sentence being parsed, we obtain its output (top-level) contextualized embedding $e_i$ from \greekbert. We then compute the following four variants of $e_i$. The $W^{(\dots)}$ matrices are learnt during fine-tuning.

\vspace*{-3mm}
\small{$$h_i^{(\text{arc-head})} = W^{(\text{arc-head})}e_i, \, h_i^{(\text{arc-dep})} = W^{(\text{arc-dep})}e_i$$
$$h_i^{(\text{rel-head})} = W^{(\text{rel-head})}e_i, \, h_i^{(\text{rel-dep})} = W^{(\text{rel-dep})}e_i$$
}
\normalsize 

\noindent $h_i^{(\text{arc-head})}$, $h_i^{(\text{arc-dep})}$ represent the $i$-th word of the sentence when considered as the head or dependent (child) of a dependency relation, respectively. $h_i^{(\text{rel-head})}$, $h_i^{(\text{rel-dep})}$ are similar, but they are used when predicting the type of a relation (see below). 

Each candidate arc from head word $j$ to dependent word $i$ is scored using the following formula, where $W^{(\text{arc})}$ is a learnt biaffine attention layer, and $b^{(\text{arc})}$ is a learnt bias capturing the fact that some words tend to be used more (or less) often as heads. 

\small
$$
s_{ij}^{(\text{arc})} = (h_j^{(\text{arc-head})})^T W^{(\text{arc})} h_i^{(\text{arc-dep})} + (h_j^{(\text{arc-head})})^T b^{(\text{arc})} $$
\normalsize

\noindent At inference time, for each word $i$, we greedily select its (unique) most probable head $y_i^{(\text{arc})}$.\footnote{We leave for future work the possibility of adding a non-greedy decoder, e.g., based on the work of \citet{chu/liu} and \citet{edmonds}, which would also guarantee that the output is always a tree.}

\small
$$ y_i^{(\text{arc})} = \arg\max_j s_{ij}^{(\text{arc})}$$
\normalsize

\noindent During training, we minimize the categorical cross entropy loss of $y_i^{(\text{arc})}$, where the possible values of $y_i^{(\text{arc})}$ correspond to the other words of the sentence. 

For a given arc from head word $j$ to dependent word $i$, its candidate labels $k$ are scored as follows, where $\oplus$ denotes vector concatenation.

\scriptsize 
$$ s_{ijk}^{(\text{rel})} = (h_j^{(\text{rel-head})})^T U_k^{(\text{rel})} h_i^{(\text{rel-dep})} + w_k^T (h_i^{(\text{rel-head})} \oplus h_i^{(\text{rel-dep})}) + b_k^{(\text{rel})}$$\normalsize

\noindent Here $U_k^{(\text{rel})}$ is a learnt biaffine layer, different per label  $k$, whereas $w_k^T$ is a learnt vector that in effect scores separately the head and the dependent word, and $b_k^{(rel)}$ is the bias of label $k$. At inference time, having first greedily selected the head $y_i^{(\text{arc})}$ of each dependent word $i$, we then greedily select the label of the arc as follows.

\small
$$ y_i^{(\text{rel})} = \arg\max_k s_{iy_i^{(\text{arc})}k}^{(\text{rel})}$$
\normalsize

\noindent During training, we minimize the categorical cross-entropy loss of $y_i^{(\text{rel})}$.
The arc prediction and label prediction components are trained jointly, adding the two cross entropy losses. 

The parser was trained and evaluated on the same modern Greek part of the Universal Dependencies dataset of Section~\ref{subsec:pos_tagging}, now using the dependency relation annotations. Consult \citet{Dikonimaki2021} and \citet{Kyriakakis2018} for more details.

Table~\ref{tab:dp_model_scores} evaluates the dependency parser of \greeknlptoolkit against those of \spacy and \stanza, using  Unlabeled Attachment Score (\textsc{uas}) and Labeled Attachment Score (\textsc{las}) on the test subset. \textsc{uas} is the percentage of the sentence's words that get the correct head, while  \textsc{las} is the percentage of words that get both the correct head and label. \greeknlptoolkit clearly provides state-of-the-art performance for this task too.

\begin{table}[h]
\small
\centering
\begin{tabular}{l|ccc}
\hline
\textbf{Score} & \textbf{\spacy} & \textbf{\stanza} & \textbf{\greeknlptoolkit} \\
\hline
\textsc{uas} & 0.66 & 0.91 & \textbf{0.94}\\
\textsc{las} & 0.64 & 0.88 & \textbf{0.92}\\
\hline
\end{tabular}
\caption{Test \textsc{uas} and \textsc{las} scores (dependency parsing).}
\label{tab:dp_model_scores}
\end{table}\vspace{-4mm}


\subsection{Greeklish-to-Greek transliteration}\label{subsec:g2g}


For Greeklish-to-Greek, we reproduced the \bytfive model of \citet{toumazatos-etal-2024-still-all-greeklish-to-me}, which was the best one, excluding \gptfour. \bytfive \cite{xue-etal-2022-byt5} operates directly on bytes, making it particularly well-suited for tasks involving text written in multiple alphabets (Greek and Latin in our case). \citet{toumazatos-etal-2024-still-all-greeklish-to-me} fine-tuned \bytfive especially for Greeklish-to-Greek, using synthetic data. The model was then evaluated on both synthetic and real-life Greeklish. Consult \citet{toumazatos-etal-2024-still-all-greeklish-to-me} for more details and evaluation results. Recall that no other modern Greek toolkit currently supports Greeklish-to-Greek (Table~\ref{tab:related_nlp_toolkits}). 

A limitation of the Greeklish-to-Greek model included in \greeknlptoolkit is that it has not been trained on Greeklish that also includes English (code switching), which is a common phenomenon in online modern Greek. This is a limitation inherited from the work of \citet{toumazatos-etal-2024-still-all-greeklish-to-me}. We are currently working on an improved Greeklish-to-Greek model that will also handle code switching. We are also considering including in \greeknlptoolkit an older statistical Greeklish-to-Greek model \cite{chalamandaris-etal-2006-allgreektome-greeklish}, which still performed well in the experiments of \citet{toumazatos-etal-2024-still-all-greeklish-to-me} and can already handle code-switching.


\section{The \greeknlptoolkit demo space}
\label{sec:the_toolkit_demo}

\begin{figure}[h]
  \includegraphics[width=\columnwidth]{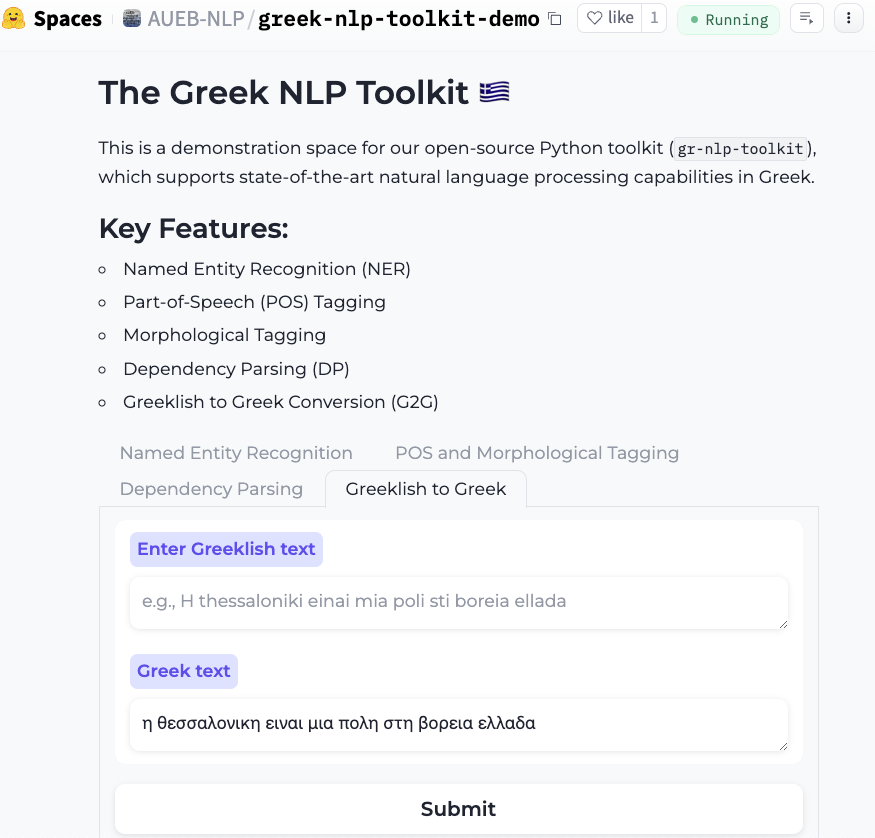}
  \caption{Example of \greeknlptoolkit's demonstration space at \url{https://huggingface.co/spaces/AUEB-NLP/greek-nlp-toolkit-demo}. The example shows Greeklish-to-Greek transliteration, but the demo provides access to the other functionalities too (\pos and morphological tagging, dependency parsing, \ner).}
  \label{fig:demo_example_g2g}
\end{figure}

For users wishing to explore \greeknlptoolkit instantly, in a no-code fashion, we also developed a demonstration space, which is open access and hosted at \url{https://huggingface.co/spaces/AUEB-NLP/greek-nlp-toolkit-demo}. Users can select tasks (\pos and morphological tagging, \ner, dependency parsing, Greeklish-to-Greek), submit their input and see the results in the user interface. 
Figure~\ref{fig:demo_example_g2g} shows an example of Greeklish-to-Greek.

\section{The \greeknlpapi}
\label{sec:the_greek_nlp_api}

Based on \greeknlptoolkit, we also developed a publicly available \api (with the same non-commercial license). The \api is hosted at \url{https://huggingface.co/spaces/AUEB-NLP/The-Greek-NLP-API}. 
It is intended to be used in research and educational applications, even applications not developed in Python, via \httpapi calls and exchange of \textsc{json} objects. \greeknlpapi conforms to the \textsc{openapi} standards.\footnote{\url{https://www.openapis.org/}}

\section{Conclusions}\label{conclusion}

We introduced \greeknlptoolkit, an open-source \nlp toolkit with state-of-the-art performance for modern Greek. It can be easily installed in Python (\mbox{\texttt{pip install gr-nlp-toolkit}}), and its code is available on Github (\url{https://github.com/nlpaueb/gr-nlp-toolkit/}). 

The toolkit currently supports \pos and morphological tagging, dependency parsing, named entity recognition, and Greeklish-to-Greek transliteration. We also presented an interactive no-code demonstration space that provides the full functionality of the toolkit (\url{https://huggingface.co/spaces/AUEB-NLP/greek-nlp-toolkit-demo}), as well as a publicly available \api at \url{https://huggingface.co/spaces/AUEB-NLP/The-Greek-NLP-API}, which allows using the toolkit even in applications not developed in Python. We discussed the methods that power the toolkit under the hood, and reported experimental results against \spacy and \stanza. 

In future work, we plan to add more tools, e.g., for toxicity detection and sentiment analysis. We welcome open-source collaboration.


\section*{Acknowledgments}

This work has been partially supported by project MIS 5154714 of the National Recovery and Resilience Plan Greece 2.0 funded by the European Union under the NextGenerationEU Program. Also, a significant portion of this work was also conducted as part of the Google Summer of Code (2024) program with the Open Technologies Alliance (GFOSS) (\url{https://eellak.ellak.gr/}). Lastly, we would like to sincerely
thank the Hellenic Artificial Intelligence Society (EETN) (\url{https://www.eetn.gr/en/)} for their
sponsorship.

\bibliography{coling_latex}

\begin{thebibliography}{30}
\providecommand{\natexlab}[1]{#1}

\bibitem[{Bakagianni et~al.(2024)Bakagianni, Pouli, Gavriilidou, and Pavlopoulos}]{bakagianni2024towards}
Juli Bakagianni, Kanella Pouli, Maria Gavriilidou, and John Pavlopoulos. 2024.
\newblock Towards {S}ystematic {M}onolingual {NLP} surveys: {G}en{A} of {G}reek {NLP}.
\newblock \emph{arXiv preprint arXiv:2407.09861}.

\bibitem[{Bartziokas et~al.(2020)Bartziokas, Mavropoulos, and Kotropoulos}]{bartziokas2020datasets-elner}
Nikos Bartziokas, Thanassis Mavropoulos, and Constantine Kotropoulos. 2020.
\newblock \href {https://doi.org/10.1145/3411408.3411437} {Datasets and {P}erformance {M}etrics for {G}reek {N}amed {E}ntity {R}ecognition}.
\newblock In \emph{{11th Hellenic Conference on Artificial Intelligence (SETN 2020)}}, SETN 2020, pages 160--167, New York, NY, USA. Association for Computing Machinery.

\bibitem[{Bird et~al.(2009)Bird, Klein, and Loper}]{bird2009natural-nltk}
Steven Bird, Ewan Klein, and Edward Loper. 2009.
\newblock \emph{Natural {L}anguage {P}rocessing with {P}ython: {A}nalyzing {T}ext with the {N}atural {L}anguage {T}oolkit}.
\newblock " O'Reilly Media, Inc.".

\bibitem[{Bojanowski et~al.(2017)Bojanowski, Grave, Joulin, and Mikolov}]{bojanowski-etal-2017-enriching}
Piotr Bojanowski, Edouard Grave, Armand Joulin, and Tomas Mikolov. 2017.
\newblock \href {https://doi.org/10.1162/tacl_a_00051} {Enriching {W}ord {V}ectors with {S}ubword {I}nformation}.
\newblock \emph{Transactions of the Association for Computational Linguistics}, 5:135--146.

\bibitem[{Chalamandaris et~al.(2006)Chalamandaris, Protopapas, Tsiakoulis, and Raptis}]{chalamandaris-etal-2006-allgreektome-greeklish}
Aimilios Chalamandaris, Athanassios Protopapas, Pirros Tsiakoulis, and Spyros Raptis. 2006.
\newblock \href {http://www.lrec-conf.org/proceedings/lrec2006/pdf/390_pdf.pdf} {All {G}reek to me! an automatic {G}reeklish to {G}reek transliteration system}.
\newblock In \emph{Proceedings of the Fifth International Conference on Language Resources and Evaluation ({LREC}{'}06)}, Genoa, Italy. European Language Resources Association (ELRA).

\bibitem[{Chu and Liu(1965)}]{chu/liu}
Y.-J. Chu and T.-H Liu. 1965.
\newblock On the shortest arborescence of a directed graph.
\newblock \emph{Science Sinica}, 14:1396--1400.

\bibitem[{Conneau et~al.(2020)Conneau, Khandelwal, Goyal, Chaudhary, Wenzek, Guzm{\'a}n, Grave, Ott, Zettlemoyer, and Stoyanov}]{conneau-etal-2020-unsupervised}
Alexis Conneau, Kartikay Khandelwal, Naman Goyal, Vishrav Chaudhary, Guillaume Wenzek, Francisco Guzm{\'a}n, Edouard Grave, Myle Ott, Luke Zettlemoyer, and Veselin Stoyanov. 2020.
\newblock \href {https://doi.org/10.18653/v1/2020.acl-main.747} {Unsupervised {C}ross-lingual {R}epresentation {L}earning at {S}cale}.
\newblock In \emph{Proceedings of the 58th Annual Meeting of the Association for Computational Linguistics}, pages 8440--8451, Online. Association for Computational Linguistics.

\bibitem[{Dikonimaki(2021)}]{Dikonimaki2021}
C.~Dikonimaki. 2021.
\newblock A {T}ransformer-based natural language processing toolkit for {G}reek -- {P}art of speech tagging and dependency parsing.
\newblock Technical report, BSc thesis, Department of Informatics, Athens University of Economics and Business.
\newblock \url{http://nlp.cs.aueb.gr/theses/dikonimaki_bsc_thesis.pdf}.

\bibitem[{Dozat et~al.(2017)Dozat, Qi, and Manning}]{dozat-etal-2017-stanfords}
Timothy Dozat, Peng Qi, and Christopher~D. Manning. 2017.
\newblock \href {https://doi.org/10.18653/v1/K17-3002} {{S}tanford{'}s {G}raph-based {N}eural {D}ependency {P}arser at the {C}o{NLL} 2017 shared task}.
\newblock In \emph{Proceedings of the {C}o{NLL} 2017 Shared Task: Multilingual Parsing from Raw Text to Universal Dependencies}, pages 20--30, Vancouver, Canada. Association for Computational Linguistics.

\bibitem[{Edmonds(1967)}]{edmonds}
J.~Edmonds. 1967.
\newblock Optimum branchings.
\newblock \emph{Journal of Research of the National Bureau of Standards B}, 71(4):233--240.

\bibitem[{Ghotsoulia et~al.(2007)Ghotsoulia, Desypri, Koutsombogera, Prokopidis, and Papageorgiou}]{GDKPP_2007}
Voula Ghotsoulia, Elina Desypri, Maria Koutsombogera, Prokopis Prokopidis, and Haris Papageorgiou. 2007.
\newblock Towards a {F}rame {S}emantics {R}esource for {G}reek.
\newblock In \emph{Proceedings of The Sixth Workshop on Treebanks and Linguistic Theories (TLT 2007)}, Bergen, Norway. University of Bergen.

\bibitem[{Honnibal et~al.(2020)Honnibal, Montani, Van~Landeghem, and Boyd}]{honnibal2020spacy}
Matthew Honnibal, Ines Montani, Sofie Van~Landeghem, and Adriane Boyd. 2020.
\newblock \href {https://doi.org/10.5281/zenodo.1212303} {{spaCy}: {I}ndustrial-strength {N}atural {L}anguage {P}rocessing in {P}ython}.

\bibitem[{Jiang et~al.(2023)Jiang, Sablayrolles, Mensch, Bamford, Chaplot, de~las Casas, Bressand, Lengyel, Lample, Saulnier, Lavaud, Lachaux, Stock, Scao, Lavril, Wang, Lacroix, and Sayed}]{jiang2023mistral7b}
Albert~Q. Jiang, Alexandre Sablayrolles, Arthur Mensch, Chris Bamford, Devendra~Singh Chaplot, Diego de~las Casas, Florian Bressand, Gianna Lengyel, Guillaume Lample, Lucile Saulnier, Lélio~Renard Lavaud, Marie-Anne Lachaux, Pierre Stock, Teven~Le Scao, Thibaut Lavril, Thomas Wang, Timothée Lacroix, and William~El Sayed. 2023.
\newblock \href {https://arxiv.org/abs/2310.06825} {Mistral {7B}}.
\newblock \emph{Preprint}, arXiv:2310.06825.

\bibitem[{Koutsikakis et~al.(2020)Koutsikakis, Chalkidis, Malakasiotis, and Androutsopoulos}]{greekbert20020}
John Koutsikakis, Ilias Chalkidis, Prodromos Malakasiotis, and Ion Androutsopoulos. 2020.
\newblock \href {https://doi.org/10.1145/3411408.3411440} {{GREEK-BERT}: {T}he {G}reeks {V}isiting {S}esame {S}treet}.
\newblock In \emph{11th {H}ellenic {C}onference on {A}rtificial {I}ntelligence}, SETN 2020, pages 110--117, New York, NY, USA. Association for Computing Machinery.

\bibitem[{Koutsogiannis and Mitsikopoulou(2017)}]{koutsogiannis-greeklish-2017}
Dimitris Koutsogiannis and Bessie Mitsikopoulou. 2017.
\newblock \href {https://doi.org/10.1111/j.1083-6101.2003.tb00358.x} {Greeklish and {G}reekness: {T}rends and {D}iscourses of “{G}localness”}.
\newblock \emph{Journal of Computer-Mediated Communication}, 9(1):JCMC918.

\bibitem[{Kyriakakis(2018)}]{Kyriakakis2018}
M.~Kyriakakis. 2018.
\newblock Exploring deep neural network models of syntax with a focus on {G}reek.
\newblock Technical report, MSc thesis, Department of Informatics, Athens University of Economics and Business.
\newblock \url{http://nlp.cs.aueb.gr/theses/kiriakakis_msc_thesis.pdf}.

\bibitem[{Loshchilov et~al.(2017)Loshchilov, Hutter et~al.}]{loshchilov2017fixing}
Ilya Loshchilov, Frank Hutter, et~al. 2017.
\newblock Fixing {W}eight {D}ecay {R}egularization in {A}dam.
\newblock \emph{arXiv preprint arXiv:1711.05101}, 5.

\bibitem[{Luccioni et~al.(2024)Luccioni, Jernite, and Strubell}]{power-hungry-processing-24}
Sasha Luccioni, Yacine Jernite, and Emma Strubell. 2024.
\newblock \href {https://doi.org/10.1145/3630106.3658542} {Power {H}ungry {P}rocessing: {W}atts {D}riving the {C}ost of {AI} {D}eployment?}
\newblock In \emph{Proceedings of the 2024 ACM Conference on Fairness, Accountability, and Transparency}, FAccT '24, page 85–99, New York, NY, USA. Association for Computing Machinery.

\bibitem[{Markantonatou et~al.()Markantonatou, Stamou, and Vak}]{UD_Greek_GUD-markantonatou}
Stella Markantonatou, Vivian Stamou, and Socrates Vak.
\newblock Gud {G}reek-{GUD}: {G}reek {U}niversal {D}ependencies {T}reebank.
\newblock \url{https://github.com/UniversalDependencies/UD_Greek-GUD}.

\bibitem[{Papageorgiou et~al.(2006)Papageorgiou, Desipri, Koutsombogera, Pouli, and Prokopidis}]{PDKPP_2006}
Harris Papageorgiou, Elina Desipri, Maria Koutsombogera, Kanella Pouli, and Prokopis Prokopidis. 2006.
\newblock Adding {M}ulti-layer {S}emantics to the {G}reek {D}ependency {T}reebank.
\newblock In \emph{Proceedings of The Fifth International Conference on Language and Evaluation (LREC-2006)}, Genoa, Italy. ELRA.

\bibitem[{Papantoniou and Tzitzikas(2020)}]{Papantoniou2020}
Katerina Papantoniou and Yannis Tzitzikas. 2020.
\newblock \href {https://doi.org/10.1145/3411408.3411410} {{NLP} for the {G}reek language: A brief survey}.
\newblock In \emph{11th Hellenic Conference on Artificial Intelligence}, SETN 2020, page 101–109, Athens, Greece.

\bibitem[{Prokopidis et~al.(2005)Prokopidis, Desypri, Koutsombogera, Papageorgiou, and Piperidis}]{PDKPP_2005}
Prokopis Prokopidis, Elina Desypri, Maria Koutsombogera, Haris Papageorgiou, and Stelios Piperidis. 2005.
\newblock \href {http://www.ilsp.gr/homepages/prokopidis/documents/gdt_tlt2005.pdf} {Theoretical and {P}ractical {I}ssues in the {C}onstruction of a {G}reek {D}ependency {T}reebank}.
\newblock In \emph{Proceedings of The Fourth Workshop on Treebanks and Linguistic Theories (TLT 2005)}, pages 149--160, Barcelona, Spain. Universitat de Barcelona.

\bibitem[{Prokopidis and Papageorgiou(2017)}]{greek-ud-treebank-1-prokopidis-papageorgiou:2017:UDW}
Prokopis Prokopidis and Haris Papageorgiou. 2017.
\newblock \href {http://www.aclweb.org/anthology/W17-0413.pdf} {Universal {D}ependencies for {G}reek}.
\newblock In \emph{Proceedings of the NoDaLiDa 2017 Workshop on Universal Dependencies (UDW 2017)}, pages 102--106, Gothenburg, Sweden. Association for Computational Linguistics.

\bibitem[{Prokopidis and Papageorgiou(2014)}]{greek-ud-treebank-2-prokopidis-papageorgiou:2014:SPMRL-SANCL}
Prokopis Prokopidis and Harris Papageorgiou. 2014.
\newblock \href {http://www.aclweb.org/anthology/W14-6109.pdf} {Experiments for {D}ependency {P}arsing of {G}reek}.
\newblock In \emph{Proceedings of the First Joint Workshop on Statistical Parsing of Morphologically Rich Languages and Syntactic Analysis of Non-Canonical Languages}, pages 89--96, Dublin, Ireland.

\bibitem[{Prokopidis and Piperidis(2020)}]{neural-resources-for-greek-nlp-prokopidis-and-piperidis-2020}
Prokopis Prokopidis and Stelios Piperidis. 2020.
\newblock \href {https://doi.org/10.1145/3411408.3411430} {A neural nlp toolkit for greek}.
\newblock In \emph{11th Hellenic Conference on Artificial Intelligence}, SETN 2020, page 125–128, New York, NY, USA. Association for Computing Machinery.

\bibitem[{Qi et~al.(2020)Qi, Zhang, Zhang, Bolton, and Manning}]{qi-etal-2020-stanza}
Peng Qi, Yuhao Zhang, Yuhui Zhang, Jason Bolton, and Christopher~D. Manning. 2020.
\newblock \href {https://doi.org/10.18653/v1/2020.acl-demos.14} {{S}tanza: {A} {P}ython {N}atural {L}anguage {P}rocessing {T}oolkit for {M}any {H}uman {L}anguages}.
\newblock In \emph{Proceedings of the 58th Annual Meeting of the Association for Computational Linguistics: {S}ystem {D}emonstrations}, pages 101--108, Online. Association for Computational Linguistics.

\bibitem[{Smyrnioudis(2021)}]{Smyrnioudis2021}
N.~Smyrnioudis. 2021.
\newblock A {T}ransformer-based natural language processing toolkit for {G}reek -- {N}amed entity recognition and multi-task learning.
\newblock Technical report, BSc thesis, Department of Informatics, Athens University of Economics and Business.
\newblock \url{http://nlp.cs.aueb.gr/theses/smyrnioudis_bsc_thesis.pdf}.

\bibitem[{Toumazatos et~al.(2024)Toumazatos, Pavlopoulos, Androutsopoulos, and Vassos}]{toumazatos-etal-2024-still-all-greeklish-to-me}
Anastasios Toumazatos, John Pavlopoulos, Ion Androutsopoulos, and Stavros Vassos. 2024.
\newblock \href {https://aclanthology.org/2024.lrec-main.1330} {Still {A}ll {G}reeklish to {M}e: {G}reeklish to {G}reek transliteration}.
\newblock In \emph{Proceedings of the 2024 Joint International Conference on Computational Linguistics, Language Resources and Evaluation (LREC-COLING 2024)}, pages 15309--15319, Torino, Italia. ELRA and ICCL.

\bibitem[{Voukoutis et~al.(2024)Voukoutis, Roussis, Paraskevopoulos, Sofianopoulos, Prokopidis, Papavasileiou, Katsamanis, Piperidis, and Katsouros}]{voukoutis2024meltemiopenlargelanguage}
Leon Voukoutis, Dimitris Roussis, Georgios Paraskevopoulos, Sokratis Sofianopoulos, Prokopis Prokopidis, Vassilis Papavasileiou, Athanasios Katsamanis, Stelios Piperidis, and Vassilis Katsouros. 2024.
\newblock \href {https://arxiv.org/abs/2407.20743} {Meltemi: The first open {L}arge {L}anguage {M}odel for {G}reek}.
\newblock \emph{Preprint}, arXiv:2407.20743.

\bibitem[{Xue et~al.(2022)Xue, Barua, Constant, Al-Rfou, Narang, Kale, Roberts, and Raffel}]{xue-etal-2022-byt5}
Linting Xue, Aditya Barua, Noah Constant, Rami Al-Rfou, Sharan Narang, Mihir Kale, Adam Roberts, and Colin Raffel. 2022.
\newblock \href {https://doi.org/10.1162/tacl_a_00461} {{B}y{T}5: Towards a {T}oken-{F}ree {F}uture with {P}re-trained {B}yte-to-{B}yte {M}odels}.
\newblock \emph{Transactions of the Association for Computational Linguistics}, 10:291--306.

\end{thebibliography}

\appendix

\section{Appendix}\label{sec:appendix}

\subsection{Hyperparameter tuning}

Table \ref{tab:greekbert_based_features_hyperparameters} provides information on the hyperparameters of the models we use for \ner, \pos tagging, morphological tagging, and dependency parsing.

\begin{table}[h!]
\centering
\begin{tabular}{|l|l|}
\hline
\textbf{Hyperparameter}              & \textbf{Range}                  \\ \hline
Learning rate                   & [5e-5, 3e-5, 2e-5]              \\ \hline
Dropout                         & [0, 0.1, 0.2]                   \\ \hline
Grad accumulation steps         & [4, 8]                          \\ \hline
Weight decay ($\lambda$)       & [0.2, 0.5, 0.8]                 \\ \hline
\end{tabular}
\caption{Hyperparameter space 
of the \ner, \pos tagging, morphological tagging, and dependency parsing models.}
\label{tab:greekbert_based_features_hyperparameters}
\end{table}

\subsection{List of Contributions\footnote{We follow the Contributor Role Taxonomy (CRediT). Consult \href{http://www.credit.niso.org}{http://www.credit.niso.org}.}
}


\hspace*{1em}\textbf{Lefteris Loukas:} Conceptualization, Software, Project administration, Funding acquisition, Writing. Lefteris led the software's refactoring to the current version, after identifying limitations in the first early one. He secured funding, supervised the development of the revamped toolkit, and created the demonstration space as well as the \api. He also co-authored this publication.

\textbf{Nikolaos Smyrnioudis:} Methodology, Formal Analysis, Software, Writing.
Nikolaos researched and created the \ner methodology, and co-developed the first version of the toolkit. Consult \citet{Smyrnioudis2021} for more information on his work, which is also summarized in \S\ref{ner_subsection}.

\textbf{Chrysa Dikonimaki:} Methodology, Formal Analysis, Software, Writing.
Chrysa researched and created the \dparsing, \pos, and morphological tagging methodologies, and co-developed the first version of the toolkit. Consult \citet{Dikonimaki2021} for more information on her work, which is also summarized in \S\ref{subsec:pos_tagging} and \S\ref{subsec:dependency_parsing}.

\textbf{Spyros Barbakos:} Software, Resources, Methodology. Spyros  refactored the previous version of the toolkit as a participant in Google's Summer of Code 2024, and enhanced it with the Greeklish-to-Greek transliteration component.

\textbf{Anastasios Toumazatos:} Software, Resources, Methodology. Anastasios provided guidance on how to integrate their Greeklish-to-Greek transliteration algorithm \cite{toumazatos-etal-2024-still-all-greeklish-to-me} in the revamped introduced toolkit.

\textbf{John Koutsikakis:} Supervision, Software, Resources. John co-supervised the BSc theses of \citet{Dikonimaki2021} and \citet{Smyrnioudis2021}, and assisted in their software and resources.

\textbf{Manolis Kyriakakis:} Software, Resources, Methodology. Manolis assisted in the development of the dependency parsing functionality, which was based on his MSc thesis \cite{Kyriakakis2018}.

\textbf{Mary Georgiou:} Software, Resources. Mary assisted in debugging and making pip-installable the first (older) version of the toolkit.

\textbf{Stavros Vassos:} Resources, Supervision. Stavros identified  current limitations in Greek NLP and provided resources for the work on Greeklish-to-Greek of \citet{toumazatos-etal-2024-still-all-greeklish-to-me}, as well as for this work.

\textbf{John Pavlopoulos:} Supervision, Writing, Methodology. John co-supervised the work on Greeklish-to-Greek of \citet{toumazatos-etal-2024-still-all-greeklish-to-me}, and this work. He co-authored this publication.

\textbf{Ion Androutsopoulos:} Supervision, Writing, Methodology. Ion co-supervised the BSc theses of \citet{Dikonimaki2021} and \citet{Smyrnioudis2021}, the Greeklish-to-Greek work of \citet{toumazatos-etal-2024-still-all-greeklish-to-me}, this work, and co-authored this publication.


\end{document}